\documentclass[runningheads]{llncs}
\usepackage{graphicx}
\usepackage{comment}
\usepackage{amsmath,amssymb} %
\usepackage{color}
\usepackage{xcolor}
\usepackage{booktabs}
\usepackage{makecell}
\usepackage{subcaption}
\usepackage[colorlinks=true]{hyperref}
\usepackage{bm}
\usepackage{xspace}
\usepackage{dsfont}
\usepackage[frozencache]{minted}
\usepackage{graphics}

\usepackage{textcomp}

\usepackage[width=122mm,left=12mm,paperwidth=146mm,height=193mm,top=12mm,paperheight=217mm]{geometry}

\makeatletter
\renewcommand\subsubsection{\@startsection{subsubsection}{4}{\z@}%
  {.5em \@plus1ex \@minus.1ex}%
  {-.5em}%
  {\normalfont\normalsize\bfseries}}
\makeatother

\include{math_commands}

\makeatletter
\DeclareRobustCommand\onedot{\futurelet\@let@token\@onedot}
\def\@onedot{\ifx\@let@token.\else.\null\fi\xspace}

\makeatother

\makeatletter
\newcommand{\printfnsymbol}[1]{%
  \textsuperscript{\@fnsymbol{#1}}%
}
\makeatother

\begin{document}
\pagestyle{headings}
\mainmatter

\title{RFUAV: A Benchmark Dataset for Unmanned Aerial Vehicle Detection and Identification}

\titlerunning{RFUAV}

\authorrunning{Shi et al.}

\author{Rui Shi \inst{1} \and Xiaodong Yu\inst{1} \and Shengming Wang \inst{1} \and Yijia Zhang \inst{1*} \and Lu Xu \inst{1*} \and Peng Pan \inst{2} \and Chunlai Ma \inst{3}}

\institute{School of Information Science and Engineering, Zhejiang Sci-Tech University, Hangzhou, Zhejiang 310018, China \email{\{rui shi, xiaodong yu, shengming wang, yijia zhang, lu xu\}xulu@zstu.edu.cn}
\and 
School of Communication and Information Engineering, Hangzhou Dianzi University, Hangzhou, Zhejiang 310018, China \email{\{peng pan\}panpeng@hdu.edu.cn}
\and
College of Electronic Engineering, National University of Defense Technology, Hefei, Anhui 230031, China \email{\{chunlai ma\}machunlai17@nudt.edu.cn}}

\maketitle

\begin{abstract}

  In this paper, we propose RFUAV as a new benchmark dataset for radio-frequency based (RF-based) unmanned aerial vehicle (UAV) identification and address the following challenges: Firstly, many existing datasets feature a restricted variety of drone types and insufficient volumes of raw data,  which fail to meet the demands of practical applications. Secondly, existing datasets often lack raw data covering a broad range of signal-to-noise ratios (SNR), or do not provide tools for transforming raw data to different SNR levels. This limitation undermines the validity of model training and evaluation. Lastly, many existing datasets do not offer open-access evaluation tools, leading to a lack of unified evaluation standards in current research within this field. RFUAV comprises approximately 1.3 TB of raw frequency data collected from 37 distinct UAVs using the Universal Software Radio Peripheral (USRP) device in real-world environments. Through in-depth analysis of the RF data in RFUAV, we define a drone feature sequence called RF drone fingerprint, which aids in distinguishing drone signals. In addition to the dataset, RFUAV provides a baseline preprocessing method and model evaluation tools. Rigorous experiments demonstrate that these preprocessing methods achieve state-of-the-art (SOTA) performance using the provided evaluation tools. The RFUAV dataset and baseline implementation are publicly available at \url{https://github.com/kitoweeknd/RFUAV/}.
  
\end{abstract}
%%%%%%%%%%%%%%%%%%
\section{Introduction}\label{sec:introduction}
In recent years, advancements in unmanned aerial vehicle (UAV) technology have significantly contributed to the widespread proliferation of UAV-related products in the market. On the one hand, this rapid development has catalyzed the swift adoption of UAVs across diverse industries and has profoundly influenced various facets of daily life. On the other hand, innovative technological advancements frequently give rise to privacy and security challenges. The proliferation of intelligent autonomous systems (IASs)~\cite{david2022resilient}, particularly UAVs used in various applications within smart cities, currently sparks concerns regarding the reliability of the underlying technologies that underpin these systems.

To address this issue, researchers have proposed various solutions, including vision-based detection systems~\cite{Visual1},~\cite{Visual2}, acoustic-based detection systems~\cite{Aco1},~\cite{Aco2}, radar-based detection systems~\cite{Radar1},~\cite{Radar2}, and radio-frequency based (RF-based) detection systems. Among these, the RF-based detection system is particularly noteworthy for its strong robustness and cost-effectiveness, making it more suitable for various application scenarios.
\section{RF-based Drone Detection System}\label{sec2}
The RF-based system leverages the inherent characteristic of drones maintaining continuous communication with their controllers throughout flight. This method uses a receiver for signal acquisition within the industrial, scientific, and medical (ISM) bands (985 MHz, 2.4 GHz, 5.8 GHz)~\cite{RFDroneViT}, which are commonly employed in drone-controller communication. The collected data then undergoes a series of signal processing techniques to detect the presence of drones.

Traditional RF-based drone detection systems are designed to identify specific models of drones within a fixed area using some signal decoding methods~\cite{dji_aeroscope},~\cite{airwarden2022}, ~\cite{schiller2023drone}. After integrating deep learning technology, this system achieves a higher degree of automation and detection accuracy. This system automatically extracts features from raw signals to accurately determine the type of drones. However, directly extracting features from data packets poses significant challenges. A common approach to address this issue is to visualize the inherent features within the data packets. Common visualization methods in signal processing include power spectral density (PSD)~\cite{PSD}, spectrogram~\cite{spectrom}, scalogram~\cite{scalogram}, and Wigner-Ville distribution~\cite{Wigner}.

\begin{figure}[h]
    \centering\small
    \includegraphics[width=1.0\columnwidth]{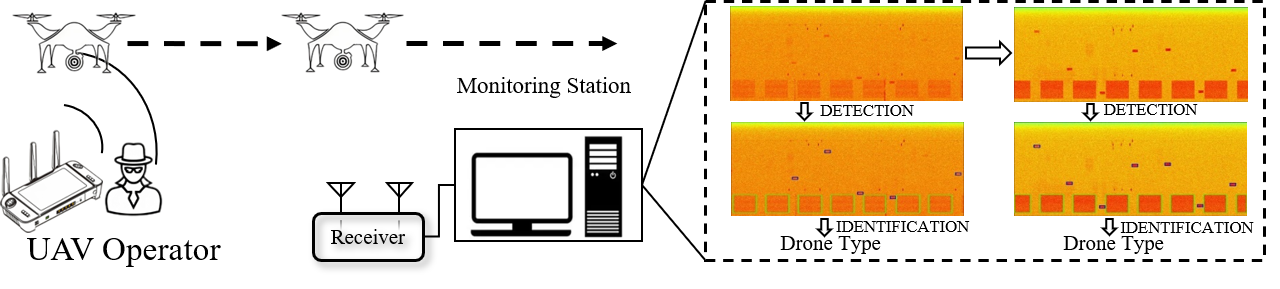}
    \caption{
    Schematic representation of the RF-based UAV monitoring framework. The system captures and analyzes radio frequency signals emitted by UAV operators to detect and classify drone activity. The monitoring station processes the received signals, facilitating UAV detection and identification of drone types. %
    }
    \label{fig_sen}
\end{figure}

In this paper, we utilize the Short-Time Fourier Transformation(STFT) to convert the raw signal into the spectrogram (for a detailed transformation procedure see Appendix. A), which visually represents the energy characteristics of the signal over time. By applying deep learning techniques to analyze these spectrograms, valuable information regarding the drone can be extracted. The entire process is illustrated in Fig.~\ref{fig_sen}. This system exploits the deep learning model’s capability to identify patterns and features within the frequency domain, thereby enabling the detection and classification of different drone types based on their unique spectral signatures.

\section{Limitations of Existing Datasets}
The existing dataset, DroneRF~\cite{DroneRF}, includes only three types of drones, whereas the market offers over 60 distinct models. Models trained on such datasets have exhibited restricted practical utility. Another dataset, DroneRFa~\cite{DroneRFa}, is significantly constrained by the signal-to-noise ratio (SNR) levels of the raw signals it contains. Although DroneRFa encompasses approximately 25 drone types, it does not provide precise SNR values for each raw data packet.

Other drone detection methods, such as vision-based systems~\cite{C2FDrone},~\cite{Using}, have gradually matured. Vision-based detection models now benefit from available benchmark dataset~\cite{Detection}, enabling fair and comprehensive evaluation of model performance. In contrast, RF-based drone detection methods lack widely accepted baseline models and benchmark dataset. Therefore, this research aims to establish a benchmark dataset for RF-based drone classification.

To provide a highly valuable public dataset, we have systematically collected raw data from 37 distinct types of UAVs under high SNR conditions in real-world environment. Our key contributions are summarized as follows:

(i) Development of a benchmark dataset for deep learning and signal processing: This research introduces a meticulously curated dataset encompassing a broad spectrum of SNRs and provides tools to adjust raw data to different SNR levels. Such capabilities are essential for conducting more realistic model evaluations. Additionally, this dataset can be utilized to assess other signal processing techniques, including decoding and drone vulnerability analysis.

(ii) A two-stage model for drone detection and identification: We present an open-source RF-based drone detection and identification architecture that leverages mainstream models such as ViT~\cite{ViT}, SwinTransformer~\cite{SwinTransformer}, YOLO~\cite{YOLO}, ResNet~\cite{ResNet}, and Faster R-CNN~\cite{ren2016fasterrcnnrealtimeobject}, EfficientNet~\cite{tan2020efficientnetrethinkingmodelscaling}, and MobileNet~\cite{MobileNet}.

(iii) Several signal processing techniques for analyzing frequency raw data: This research presents fundamental signal processing methods for drone data analysis, including SNR estimation, 3D spectrogram analysis, and decomposition of raw signal data components.
\section{RFUAV Dataset}
This section will address the following aspects: (i) the methodology for collecting raw data from UAVs, (ii) the definition of the \textit{"RF drone fingerprint"} through in-depth analysis of the raw data, (iii) the development of a two-stage model for drone detection and identification based on RFUAV, and (iv) the evaluation of several signal processing algorithms on the RFUAV.

\subsection{Data Collection}
\label{sec:set_loss}
To obtain a substantial volume of high-quality raw data, we develop a comprehensive signal acquisition platform. The platform consists of three key components: (i) a signal acquisition system for collecting raw data, (ii) a data storage system for archiving and sharing the collected data, and (iii) a real-time playback platform for inspecting the quality of acquired signals.

The hardware utilized in this setup primarily consists of the following: (i) two Universal Software Radio Peripheral (USRP) X310 units for signal acquisition and transmission, (ii) a computer equipped with an AMD Ryzen 5950X CPU, 32 GB of RAM, and a 2TB Samsung M.2 Solid State Drive (SSD), running Ubuntu 22.04 Linux and GNU Radio v3.9.10, serving as the signal collector, (iii) a HUAWEI SP310 10GbE card acting as the intermediary connecting the USRP X310 units and the signal collector, (iv) a spectrum analyzer with a frequency range of 9 kHz to 13.2 GHz, used to verify the feasibility of the acquired signals, and (v) a laptop running adobe audition (AU) for signal observation. Initially, two USRP X310 units are interfaced with the signal collector via 10GbE I/O ports. The first USRP X310 (USRPI) captures raw data, which is subsequently stored directly on the signal collector’s SSD. The second USRP X310 (USRPO) interfaces with a spectrum analyzer to assess the quality of the collected data.

\begin{figure}[h]
    \centering\small
    \includegraphics[width=1.0\columnwidth]{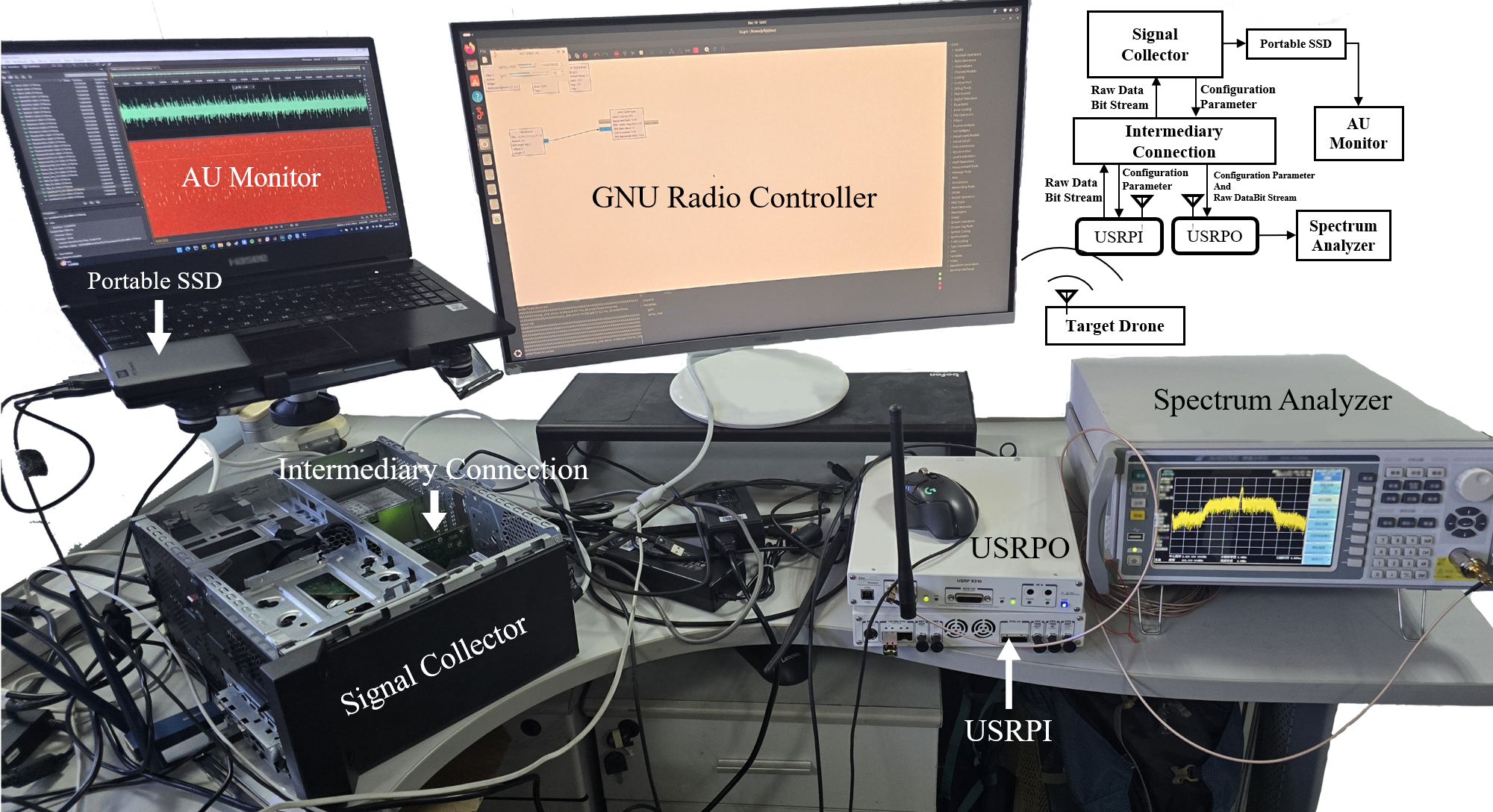}
    \caption{
    Hardware architecture and interconnections of the RF signal acquisition platform. The system integrates multiple components, including a GNU Radio controller, USRP devices, a spectrum analyzer, and an AU monitoring unit. These components facilitate high-quality RF signal collection, processing, and analysis. %
    }
    \label{fig_setup}
\end{figure}

The data collection location is carefully selected to minimize signal interference. During the collection process, the transmitters of 37 different UAV types are either directly connected to the receiver of the USRPI or positioned in close proximity to it, ensuring optimal signal quality of the collected data. Upon completion of data acquisition for a specific drone type, the collected data is transferred from the signal collector to the portable SSD. Subsequently, the laptop is used to analyze the spectrogram and PSD to verify the signal has been fully captured. A dedicated signal playback module, implemented using GNU Radio, enables USRPO to transmit the collected data packets. USRPO is directly connected to a spectrum analyzer, which facilitates the evaluation of each raw data packet's bandwidth and SNR. This systematic workflow for data collection ensures the quality of every signal and establishes a robust foundation for subsequent research (for more detailed signal collection, see Appendix. B). The schematic of the entire platform setup is illustrated in Fig.~\ref{fig_setup}.

In total, approximately 1.3 TB of raw data are collected from 37 distinct types of drones. These raw signals encompass Frequency-Hopping Spread Spectrum (FHSS) signals transmitted by the remote control to regulate the drone’s behavior. FHSS signals are characterized by a fixed hopping bandwidth (FHSBW), hopping duration (FHSDT), hopping duty cycle (FHSDC), and hopping pattern period (FHSPP). Additionally, some drones equipped with cameras possess the capability to capture and transmit images. The raw signals from these drones also include fixed bandwidth signals used for image transmission. Typically, these image transmission signals manifest near the frequency bands associated with the FHSS signals.

\subsection{Data Analysis}
\subsubsection{Basic Information about Dataset}
In RFUAV, the raw data is stored in accordance with the standard in-phase and quadrature components (IQ) sampling method~\cite{IQ-demodulation}. Each data packet is organized as interleaved IQ binary bitstreams, where each sample is represented by two 32-bit floating-point numbers (fp32). To load the data into memory, any binary file processing workflow can be applied, provided the fp32 input format is specified. The I and Q components can be independently indexed into separate vectors, which are subsequently combined as \(I+j\times Q\) to reconstruct the complete sampled signal. Furthermore, the parameters of the USRP configured during data acquisition for each drone type, are documented in a corresponding XML file. These parameter files are stored alongside the raw data to ensure seamless access to the original acquisition settings for future reference. The standard raw data can be processed into a signal processing-compatible data stream using either Matlab or Python workflows, with relevant preprocessing code available in our GitHub repository.

To enhance readers’ intuitive and comprehensive understanding of the RFUAV, we conduct an analysis of its core attributes, as illustrated in Fig.~\ref{FSM} (for more detailed results see Appendix. C). This analysis encompasses the data volume associated with each drone type, the center frequencies employed during data acquisition, and the SNR of the signals captured during the original sampling process.

\begin{figure}[h]
    \centering\small
    \includegraphics[width=1.0\columnwidth]{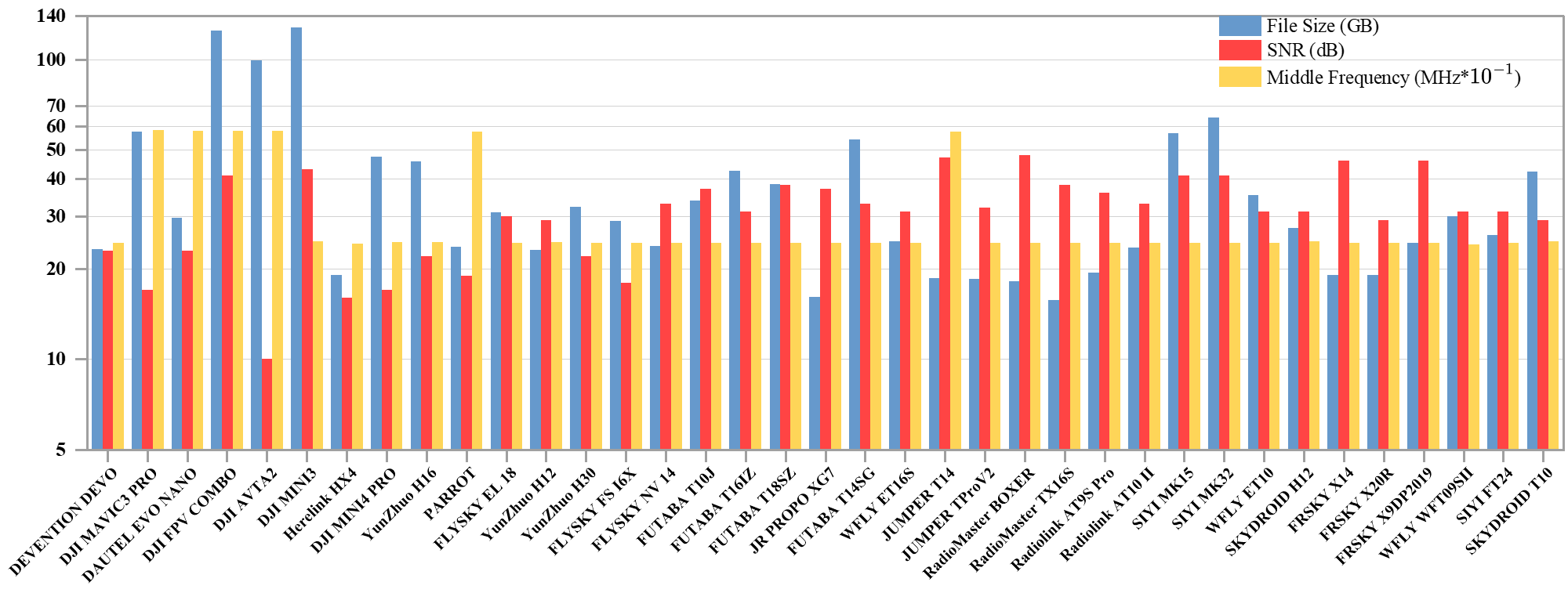}
    \caption{
    Statistical overview of the original dataset. The figure presents key attributes of the collected RF data, including file size (GB), signal-to-noise ratio (SNR) in dB, and center frequency (MHz × \(10^{-1}\)) for various UAV controllers. %
    }
    \label{FSM}
\end{figure}

As depicted in the figure, each drone type is equipped with a minimum of 15GB of raw data storage. Certain drones, such as the DJI FPV COMBO, DJI AVATA2, and DJI MINI 3, possess the capability to select various signal bandwidths for video transmission during communication. To ensure comprehensive coverage of the data with each drone, we meticulously captured all potential scenarios that may arise throughout the signal transmission process.

It is noteworthy that the frequency hopping pattern of certain drones’ FHSS signals exhibits variability contingent upon the drone’s behavior. For example, during the communication establishment process, the remote controller of the WFLY WFT09SII generates FHSS signals characterized by shorter FHSDC and irregular hopping patterns. However, once a stable connection is established between the drone and its controller, and subsequent to transmission from the flight controller to the drone, there is a transition in the FHSS signal towards a longer-period signal. At this stage, the frequency hopping pattern resembles that of a triangular waveform. To ensure comprehensive sample coverage, our dataset encompasses classifications based on various actions performed by the drone. For each individual drone, both the pairing process and flight operations are documented, thereby providing an extensive range of signal types across different operational phases.

\subsubsection{Analysis of the Main Characteristics of Drone Signal}
\label{sec:architecture}
\begin{figure}[h]
    \centering\small
    \includegraphics[width=1.0\columnwidth]{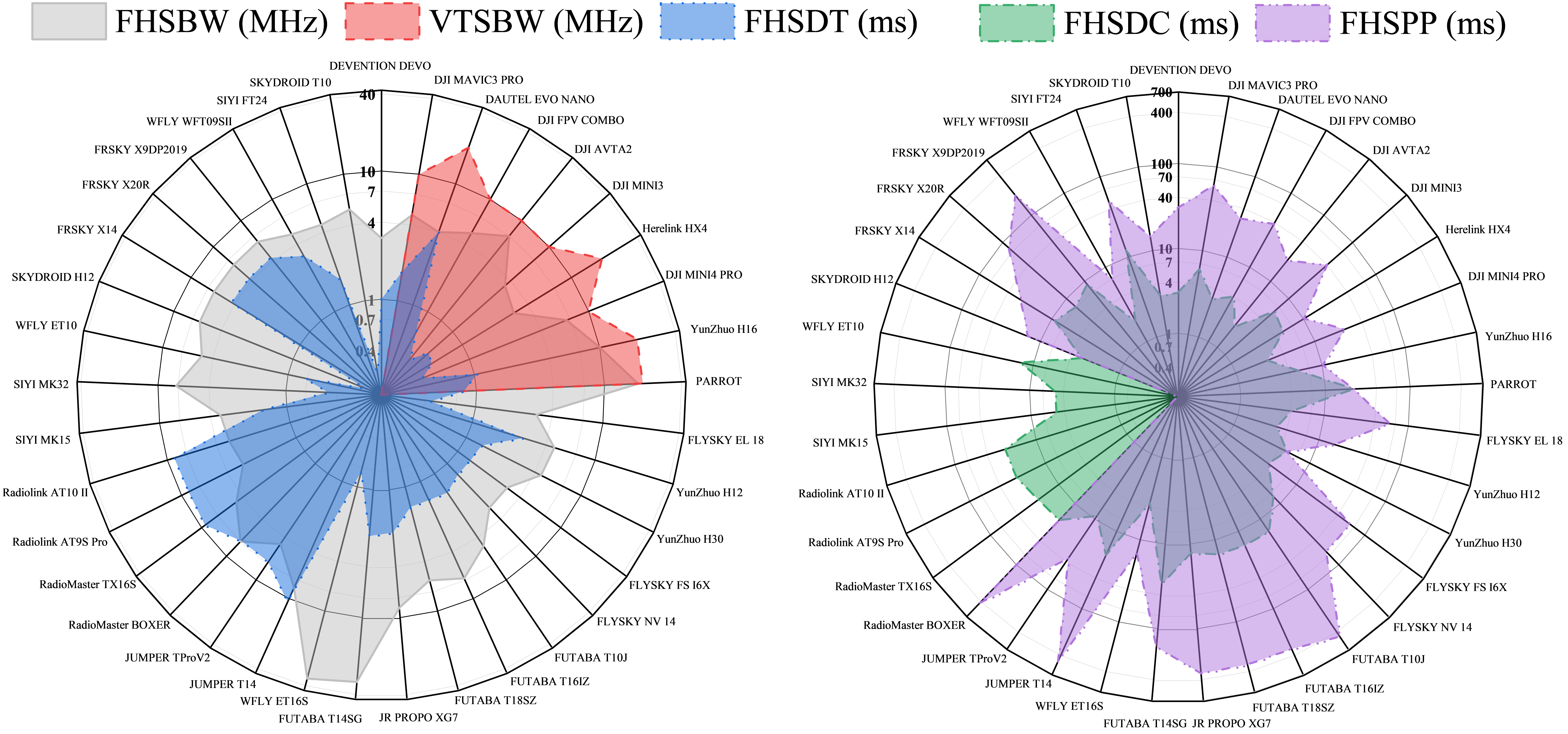}
    \caption{
    Comprehensive statistical analysis of key features observed in drone communication signals. Left Polar Plot: Displays the frequency-hopping signal bandwidth (FHSBW, MHz), image transmission signal bandwidth (VTSBW, MHz), and frequency-hopping signal duration time (FHSDT, ms). Right Polar Plot: Depicts the frequency-hopping duty cycle (FHSDC, ms) and frequency-hopping signal periodicity (FHSPP, ms).
    }
    \label{FVFFF}
\end{figure}

In the practical application of RF-based drone detection systems, signals within the same frequency band as drones also include Bluetooth, WiFi, and various other sources of interference. These interfering signals can significantly hinder the model’s capacity to accurately classify different drone types. To assist readers in identifying more effective  solutions, we conduct an analysis of the characteristics of drone signals through their time-frequency representations. Unlike signals from other sources, drone FHSS signals exhibit distinct characteristics, including fixed FHSBW and FHSDT. The FHSDC and the FHSPP vary across different drone models as shown in Fig.~\ref{FVFFF}.

As illustrated in the figure, the Herelink HX4 exhibits the smallest FHSBW at 2.9 MHz, while the WFLY-ET16S exhibits the largest FHSBW at 35.12 MHz, which is nearly the same as the video transmission signal bandwidth (VTSBW). The SKYDROID H12 shows the smallest FHSDT at 0.25 ms, whereas the JUMPER T14 exhibits the largest FHSDT at 10.73 ms. While most drones generate FHSS signals with a distinct FHSPP, several models, — including RadioMaster TX16S, Radiolink AT9S Pro, Radiolink AT10 II, SIYI MK15, SIYI MK32, and WFLY ET10 — do not exhibit a clear FHSPP within their FHSS signals. All drone-generated signals maintain a fixed FHSDC. The WFLY WFT09SII has the shortest FHSDC at 1.86 ms, while the FUTABA T14SG displays the longest FHSDC at 30.1 ms. Additionally, the YunZhuo H30 features the shortest FHSPP at 5 ms and the JUMPER T14 features the longest FHSPP at 480 ms.

By integrating the FHSBW, FHSDT, FHSDC, FHSPP, and VTSBW, a distinctive feature sequence is generated that can be likened to an \textit{"RF drone fingerprint"}. This RF drone fingerprint will serve as the primary input for deep learning networks, thereby enabling differentiation among various drone types.

\subsubsection{Floating Time-Duration of Image Transmission Signal in RFUAV}

In contrast, the video transmission signals between drones and controllers typically exhibit longer duration times and wider bandwidths when compared with the FHSS signal. These signals are characterized as fixed-frequency wideband signals, defined by predefined bandwidths and center frequencies.

\begin{figure}[h]
    \centering\small
    \includegraphics[width=1.0\columnwidth]{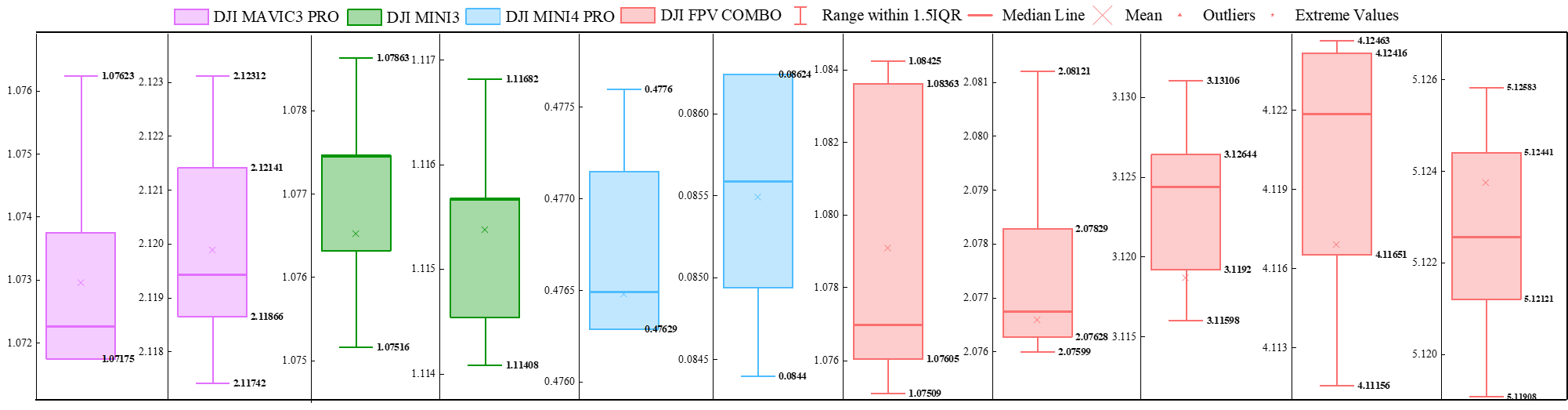}
    \caption{
     Main distribution of image transmission signal duration in the RFUAV %
    }
    \label{fig_violion}
\end{figure}

An intriguing phenomenon observed in RFUAV merits further investigation: for drones that transmit video, the duration of these image transmission signals often exhibits variability. To explore the underlying patterns governing the duration of these signals, we conduct a statistical analysis. Specifically, we measure the duration of 200 individual image transmission signals from various models, including the DJI MAVIC3 PRO, DJI MINI3, DJI MINI4 PRO, and DJI FPV COMBO. The statistical distribution derived from these measurements is illustrated in Fig.~\ref{fig_violion}.

As illustrated in the figure, the image transmission signals of these four drones display irregular variations in duration, although they remain confined within a specific range. For the DJI FPV COMBO, the signal durations exhibit five distinct values: 1.079 ms, 2.076 ms, 3.119 ms, 4.116 ms, and 5.127 ms. Within this set of values, minor fluctuations of approximately 0.015 ms are also observed in the signal durations. Similar patterns can be noted for the other three drones; however, their image transmission signals vary between only two discrete values and demonstrate smaller fluctuations.

This characteristic of temporal fluctuation is a crucial aspect that differentiates drone image transmission signals from other signal types. For example, the five distinct duration cases observed in the DJI FPV COMBO’s image transmission signals reveal notable morphological differences in their time-frequency representations. In order to execute target detection tasks effectively, these signals must be categorized into five distinct types. The minor temporal fluctuations may arise from signal transmission losses and can occasionally result in morphological variations within the signals, potentially affecting the model’s classification accuracy.

\subsection{Drone Detection and Identification Using Two-Stage Model}

To comprehensively showcase the capabilities of RFUAV, we designed a two-stage model for drone detection and identification. The first-stage model utilizes an object detection network to detect drone signals, while the second-stage model employs an image classification network for identifying the drone type.

The newly acquired radio frequency data undergoes a series of preprocessing techniques to generate spectrograms, which are then input into the first-stage network for drone signal detection. This network identifies the presence of drones within the spectrogram by detecting signal targets. The regions of the spectrogram containing detected signals are subsequently passed to the second-stage network, which classifies the drones based on their type.

In this paper, we use YOLO as the first-stage detection network and ResNet as the second-stage classification network, forming a two-stage network. During the data preprocessing phase, we employ a dual-buffer queue combined with Fast Fourier Transform (FFT) to implement an efficient time-frequency spectrogram generation algorithm. This system achieves high efficiency in drone detection tasks while maintaining high accuracy, offering strong practicality. The specific performance and implementation details can be found in our GitHub repository.

\subsection{Signal Separation and Parameter Estimation}\label{sub:signal_separation_and_parameter_estimation}
RFUAV can also be employed in various signal processing analyses. In this section, we present a signal separation and SNR estimation algorithm that utilizes a subset of signals from RFUAV.

We first employ a dual sliding window-based signal detection algorithm in the time domain to identify the starting and ending positions of a signal clip. Based on the results of this separation, we use Welch’s method to estimate the PSD of the signal and determine the center frequency of the signal. Finally, the SNR is calculated in the time domain according to the definition of SNR. The specific implementation process is detailed Appendix. D.
\section{Experiments}
\label{sec:experiments}
In our research at the intersection of signal processing and deep learning, it is essential to investigate how preset parameters for signal processing algorithms affect the accuracy of deep learning models on UAV identification tasks. This paper primarily investigates the impact of different time-frequency plot color map (CMAP) methods and frequency resolution on the performance of deep learning models on UAV identification tasks under different SNRs. We conduct two sets of experiments in Sec.~\ref{sec5.1} and Sec.~\ref{sec5.2} to determine the optimal CMAP and STFTP settings that enhance deep learning model accuracy. Through this comparative analysis, we identify the state-of-the-art (SOTA) preprocessing parameters for spectrogram perception.

\subsection{CMAP comparison}\label{sec5.1}
From the perspective of human observation, color plays a pivotal role in shaping our perception of an object's characteristics. However, within the domain of deep learning, the sensitivity of models to color variations has rarely been thoroughly studied. In this research, time-frequency spectrograms are generated by applying the STFT transformation to raw signals, with STFT values mapped to colors according to a predefined color scale.

\begin{figure}[h]
    \centering\small
    \includegraphics[width=1.0\columnwidth]{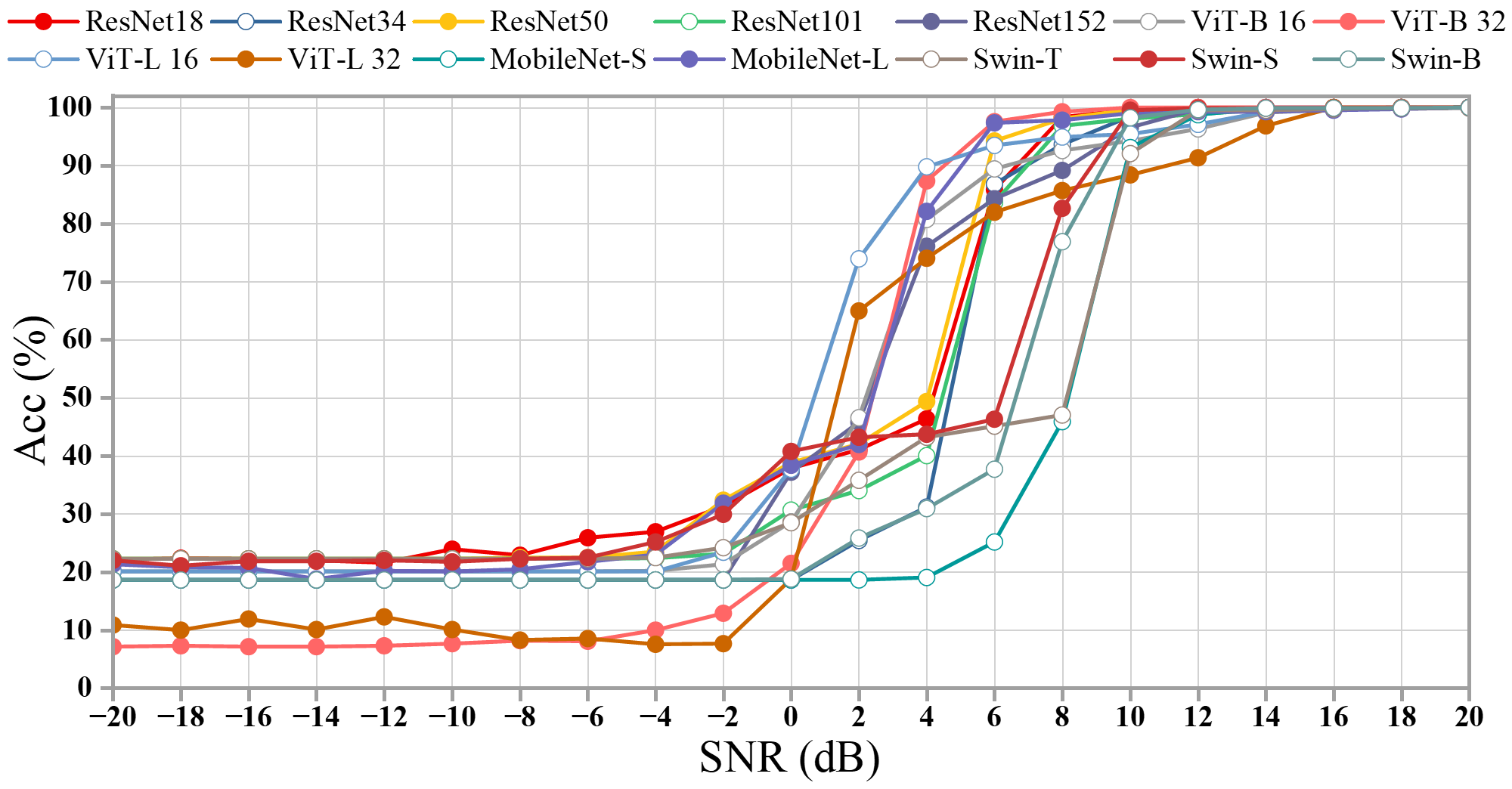}
    \caption{
    Model comparison results %
    }
    \label{res_exp1}
\end{figure}

To enhance the overall performance of models in spectrogram-based tasks and to align with the research objective of exploring model sensitivity to color variations of the same object, this study will experiment with four CMAP methods that exhibit significant visual characteristics from a human perception standpoint. Specifically, we initially fixed a single frequency resolution and experimented with four CMAP methods ("Parula", "Hot", "HSV", and "Autumn") to generate the spectrograms. A detailed portion of the spectrogram results is provided in the Appendix. E. Consequently, we perform accuracy evaluations on data with varying SNR to determine the optimal CMAP method.

\subsubsection{Dataset}\label{sec5.1.1}
Given that this experiment encompasses multiple comparative models and the primary objective of this subsection is specifically to identify the SOTA preprocessing pipeline for deep learning models, the experiments in this subsection will be conducted on a subset of the RFUAV.

We select the raw signals from five drones equipped with a camera, namely the DJI AVATA2, DJI FPV COMBO, DJI MAVIC3 PRO, DJI MINI3, and DJI MINI4 PRO. Firstly, we estimate the SNR of each raw signal based on the SNR estimation results presented in Sec.~\ref{sub:signal_separation_and_parameter_estimation}. Then, we apply additive Gaussian white noise (AWGN) to adjust the SNR of the raw signals. Finally, for each drone, we generate raw data with SNR levels ranging from -20 dB to 20 dB in 2 dB increments.

The raw signal and their SNR-adjusted counterparts are transformed into spectrograms with identical frequency resolutions, employing four distinct CMAP methods (for more detailed CMAP methods analysis, see Appendix. A) to produce images of consistent resolution. The spectrograms of the raw signals and the SNR-adjusted signals at 20 dB serve as training data, whereas the spectrograms of the SNR adjusted signals with SNR levels ranging from -20 dB to 20 dB (in 2 dB increments) are used as validation data.

\subsubsection{Model Pre-selection}\label{sec5.1.2}
Given that our task is fundamentally an image classification problem, we first conduct a model selection experiment to ensure the accuracy of the CMAP and frequency resolution comparative analyses. In this experiment, CMAP and frequency resolution are fixed to "Parula" and 256, as these settings are commonly used in signal processing. The primary objective is to identify the most suitable classification model for UAV identification from among the mainstream classification models. The selected model will then serve as the baseline model for subsequent experiments. We train ResNet, Swin Transformer, ViT, and MobileNet on the aforementioned dataset. The evaluation metrics include accuracy (Acc) and overall accuracy (OA). To facilitate a clearer understanding of Acc variations under different SNR levels, we visualize the experimental results in Fig.~\ref{res_exp1} and provide detailed numerical results in Table~\ref{exp1_table}.

\begin{table}
\begin{center}
\caption{Detailed model comparison results}
\label{exp1_table}
\begin{tabular}{c|cccc}
\hline
\makecell{Model} & \makecell{\(\text{OA}_{\text{SNR}\leq-10}(\%)\) } & \makecell{\(\text{OA}_{-8\leq \text{SNR}\leq 8}(\%)\)} & \makecell{\(\text{OA}_{\text{SNR}\geq 10}(\%)\)} & \makecell{\(\text{Acc}(\%)\)} \\
\hline
ResNet18 & \textbf{22.39} & 46.28 & 99.93 & 54.78 \\
ResNet34 & 18.67 & 36.71 & 99.65 & 49.54 \\
ResNet50 & 22.25 & 47.06 & 99.85 & 55.05 \\
ResNet101 & 22.25 & 41.69 & 99.38 & 52.61 \\
ResNet152 & 18.67 & 45.24 & 99.38 & 53.11 \\
\hline
ViT-B-16 & 20.08 & 46.62 & 98.18 & 53.77 \\
ViT-B-32 & 7.29 & 42.82 & \textbf{100.00} & 49.01 \\
ViT-L-16 & 20.08 & \textbf{52.60} & 98.55 & \textbf{56.44} \\
ViT-L-32 & 10.87 & 39.73 & 96.08 & 47.58 \\
\hline
MobileNet-S & 18.67 & 22.46 & 98.57 & 43.13 \\
MobileNet-S & 20.33 & 50.51 & 99.45 & 55.87 \\
\hline
Swintransformer-T & 22.25 & 32.32 & 98.44 & 48.34 \\
Swintransformer-S & 21.76 & 39.61 & 99.83 & 51.72 \\
Swintransformer-B & 18.67 & 29.43 & 99.57 & 46.4 \\
\hline
\end{tabular} 
\end{center}
\end{table}

As shown in Table~\ref{exp1_table}, ViT-L-16 achieve the highest OA and \(\text{OA}_{-8<\text{SNR}<8}\) at 56.44\% and 52.60\%; ViT-B-32 achieve the highest \(\text{OA}_{\text{SNR}>10}\) at 100.00\%; ResNet18 achieve the highest \(\text{OA}_{\text{SNR}<-10}\) at 22.39\%.

\subsubsection{CMAP Comparison}\label{sec5.1.3}
\begin{figure}[h]
    \centering\small
    \includegraphics[width=1.0\columnwidth]{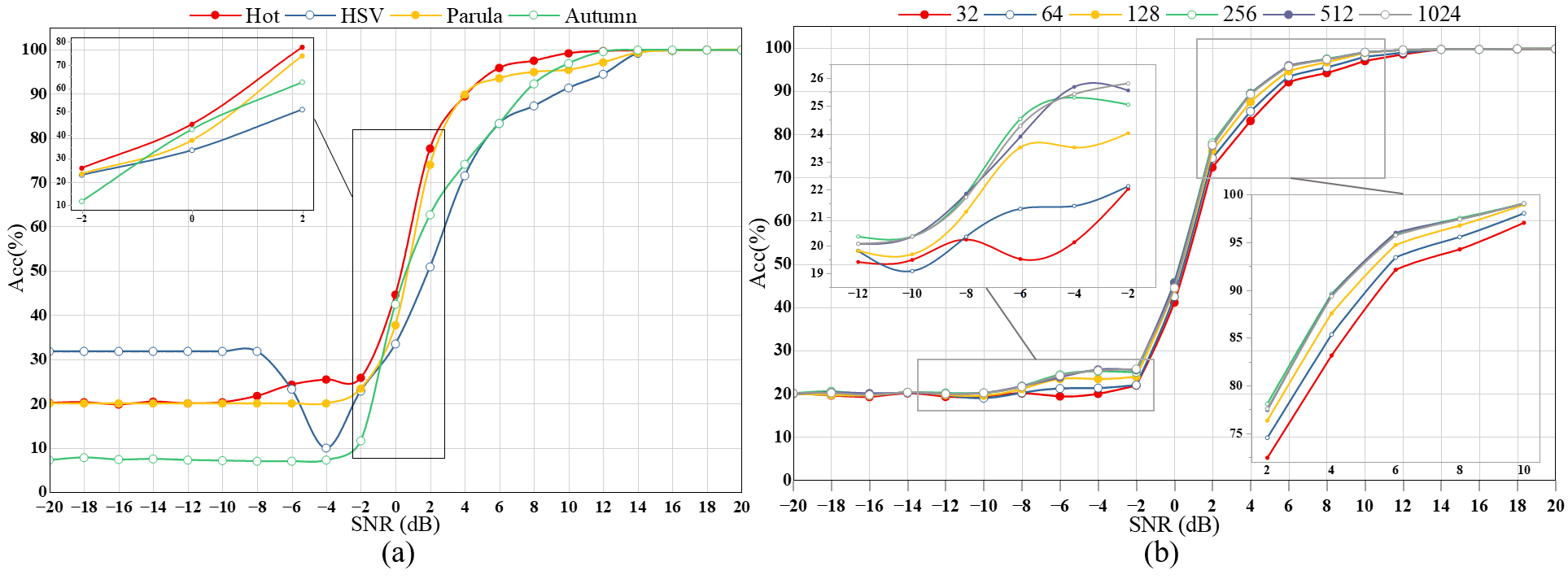}
    \caption{
    (a) presents the comparative results of CMAP. (b) presents the STFTP comparison results,%
    }
    \label{res_exp1.2}
\end{figure}

Based on the results obtained in previous experiments, we select ViT-L-16 as the baseline model for the subsequent CMAP comparison experiments. By utilizing the previously proposed dataset generation method, we construct training set and validation set of equivalent size to those produced using the "Parula", "Hot", "Autumn", and "HSV" CMAP methods. The same data augmentation techniques and model training strategies as in previous experiments are applied to train the ViT-L-16 model. Finally, the model is evaluated on their respective validation sets using the same evaluation parameters as before. The experimental results are illustrated in Fig.~\ref{res_exp1.2}, with detailed results presented in Table.~\ref{exp1.2_table}.

\begin{table}
\begin{center}
\caption{Detailed CMAP comparison results}
\label{exp1.2_table}
\begin{tabular}{c|cccc}
\hline
\makecell{CMAP Method} & \makecell{\(\text{OA}_{\text{SNR}\leq -10}(\%)\)} & \makecell{\(\text{OA}_{-8\leq \text{SNR}\leq 8}(\%)\)} & \makecell{\(\text{OA}_{\text{SNR}\geq 10}(\%)\)} & \makecell{\(\text{Acc}(\%)\)} \\
\hline
Autumn & 7.42 & 43.07 & 99.30 & 48.95 \\
\textbf{Hot} & 20.20 & \textbf{55.80} & \textbf{99.70} & \textbf{58.16} \\
Parula & 20.08 & 52.60 & 98.55 & 56.44 \\
HSV & \textbf{31.84} & 46.03 & 97.38 & 56.65 \\
\hline
\end{tabular} 
\end{center}
\end{table}

As shown in the figure and table, we observe that the ViT-L-16 model achieves the highest OA of 58.16\% on datasets using the "Hot" CMAP method. In scenarios where the SNR is between -8 dB and 20 dB, the model consistently achieves the highest precision, with \({\text{OA}}_{\text{SNR}\geq 10}\) at 99.70\% and \({\text{OA}}_{-8\le \text{SNR}\le 8}\) at 55.80\%, respectively. Meanwhile, for datasets using the "HSV" CMAP method, the ViT model achieves the highest \({\text{OA}}_{\text{SNR}\le -10}\) of 31.84\%. Therefore, the "Hot" CMAP method should be considered the most suitable option for this task.

From this experiment, we can conclude that different CMAP methods significantly impact the performance of the model, specifically under low SNR conditions. When training is conducted exclusively with high-SNR data, the model frequently fails to capture the features associated with image differences caused by SNR variations. This observation aligns with practical application scenarios and underscores the challenge deep learning aims to address generalizing solutions for broader situations using features from a limited subset of data.

\subsection{Frequency Resolution Comparison}\label{sec5.2}
The STFT transformation is used to convert a raw signal from the time domain to the frequency domain. During this process, the total number of STFT points in the discrete Fourier transform (STFTP) directly influences the accuracy of the signal's representation in the frequency domain, i.e., the frequency resolution. In this subsection, we will discuss how this parameter affects the model's accuracy.

\subsubsection{Datasets}\label{sec5.2.1}
Based on the results presented in previous experiments, we use the "Hot" as a CMAP method and apply STFT with STFTP of 32, 64, 128, 256, 512, and 1024 to the raw data from the experimental drone objects to generate spectrograms within an SNR range of -20 to 20 dB, with intervals of 2 dB, for use as the validation set. The training data consists of the raw data augmented with an SNR of 20 dB to train the model. The data augmentation methods and dataset splitting strategy for training are consistent with those outlined in Sec.~\ref{sec5.1.2}.

\subsubsection{Comparison Result}\label{sec5.2.2}

We evaluate the accuracy of the ViT-L-16 model on the aforementioned dataset, with the results presented in Table.~\ref{exp1.3_table} and Fig.~\ref{res_exp1.2}. (b).

\begin{table}
\begin{center}
\caption{Detailed STFTP comparison results}
\label{exp1.3_table}
\begin{tabular}{c|cccc}
\hline
\makecell{STFTP} & \makecell{\(\text{OA}_{\text{SNR}\leq -10}(\%)\)} & \makecell{\(\text{OA}_{-8\leq \text{SNR}\leq 8}(\%)\)} & \makecell{\(\text{OA}_{\text{SNR}\geq 10}(\%)\)} & \makecell{Acc}(\%)\\
\hline
32 & 19.76 & 51.70 & 99.14 & 56.13 \\
64 & 19.83 & 52.99 & 99.39 & 56.77 \\
128 & 19.93 & 54.70 & 99.64 & 57.60 \\
256 & \textbf{20.33} & \textbf{56.00} & 99.66 & \textbf{58.28} \\
512 & 20.25 & 55.92 & 99.66 & 58.23 \\
1024 & 20.20 & 55.80 & 99.66 & 58.16 \\
\hline
\end{tabular} 
\end{center}
\end{table}

As depicted in the table and figure, the impact of increasing STFTP on model accuracy demonstrates a boundary effect (for a more detailed analysis, refer to Appendix. D). Once STFTP reaches 256, further increments do not result in significant improvements in model accuracy. Instead, the accuracy gradually stabilizes, with larger STFTP values causing fluctuations in accuracy between 0.05\% and 0.12\%. Therefore, in this experiment, the model with STFTP set to 256, achieves the highest OA of 58.28\%, \({\text{OA}}_{\text{SNR}\le -10}\) of 20.33\%, and \({\text{OA}}_{-8\le \text{SNR}\le 8}\) of 56.00\%.

From the frequency resolution comparison experiment results, it is evident that while STFTP is a crucial parameter influencing outcomes in traditional signal processing algorithms, its impact on deep learning-based signal processing research is less pronounced. This observation highlights that in wide-spectrum signal energy analysis, raw signal analysis is not heavily dependent on frequency resolution. In practical applications, adjusting STFTP to the most suitable value based on the specific task requirements is sufficient.

\section{DISCUSSION}\label{sec6}
RFUAV exhibits significant potential for further development. The experiments presented in this paper are conducted on a subset of the RFUAV, aiming to provide valuable insights for researchers and facilitate subsequent advancements in the field. The potential of RFUAV extends beyond its current capabilities, offering extensive opportunities for enhancing detection accuracy, especially under challenging conditions such as low SNR. Future work should focus on addressing the limitations observed in our experiments, exploring novel model architectures, and developing advanced signal processing techniques to improve detection reliability and robustness across diverse operational scenarios.

The raw data preprocessing methods and deep learning model presented in this paper are also applicable to other related studies, such as DroneRFa. To further enhance the detection model, additional drone data from datasets like DroneRFa and DroneRF can be integrated with RFUAV using the same preprocessing techniques. This integration would increase the total number of drone types to over 54, thereby effectively covering nearly all commercially available civilian drone models.

To maintain the RFUAV’s relevance, we will continuously update it with raw data from newly emerging civilian drone models. This approach not only provides a foundational resource for researchers studying radio-frequency data but also supplies essential materials for advancing deep learning research in the radio-frequency domain. We anticipate that, through our ongoing efforts, this drone detection method will continue to evolve and improve.

\section{CONCLUSION}\label{sec7}

This paper introduces RFUAV as a novel benchmark dataset for RF-based drone detection and identification. The comprehensive data analysis and diverse use case examples underscore the significant potential inherent in RFUAV. Through rigorous experimentation, we have demonstrated a SOTA data preprocessing method that effectively integrates deep learning with signal processing. In future work, we plan to continuously update the dataset with data from newly emerging drone models.

\bibliography{paperRef}
\bibliographystyle{splncs04}

\appendix
\newpage
\section{STFTP Transformation Detail}\label{app2}
In this section, we demonstrate the effect of the total number of STFT points in the discrete Fourier transform (STFTP) on the quality of the spectrogram.

The STFT expression for discrete signals is shown in Eq.~\ref{eq.STFT}:
\begin{eqnarray}
\label{eq.STFT}
X[m, k] = \sum_{n=-\infty}^{\infty} x[n] w[n - m] e^{-j2\pi k n / N}
\end{eqnarray}
where, \( x[n] \) is the discrete input signal, \( w[n - m] \) is the discrete window function centered at \( m \), \( k \) represents the frequency bin index, \( N \) is the STFTP. In this paper, we use the Hamming Window as the window function. The expression for the Hamming Window is presented in Eq.~\ref{eq.hanning}:

\begin{eqnarray}
\label{eq.hanning}
w[n] = 0.54 - 0.46 \cos\left( \frac{2\pi n}{N-1} \right), \quad 0 \leq n \leq N-1
\end{eqnarray}
where \( w[n] \) is the window function at discrete time index \( n \), \( N \) indicates the STFTP.

\begin{figure}[h]
    \centering\small
    \includegraphics[width=1.0\columnwidth]{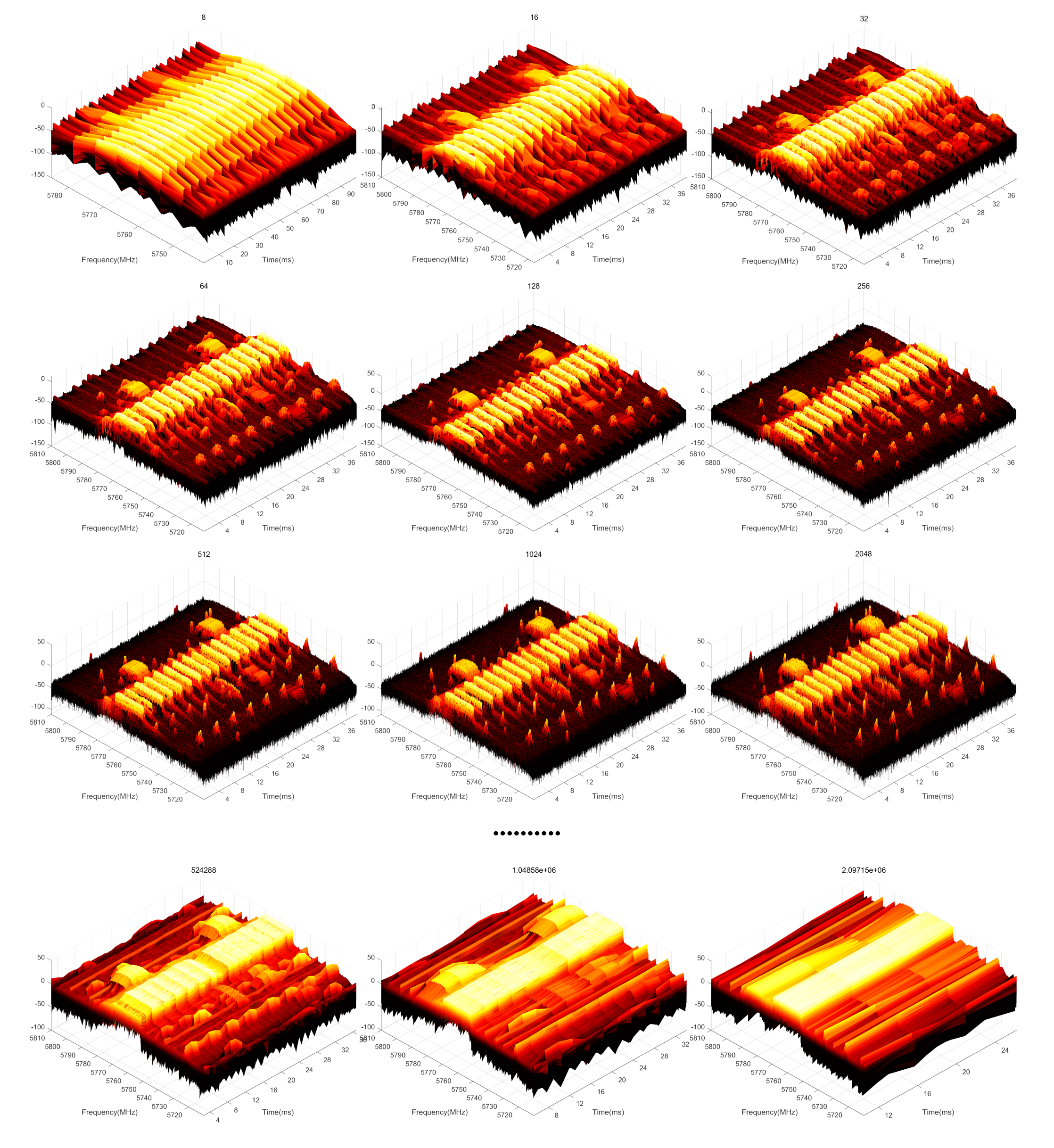}
    \caption{
    DJI MINI4 PRO 3D spectrograms under different STFTPs %
    }
    \label{fig.app2_1}
\end{figure}

\begin{figure}[h]
    \centering\small
    \includegraphics[width=1.0\columnwidth]{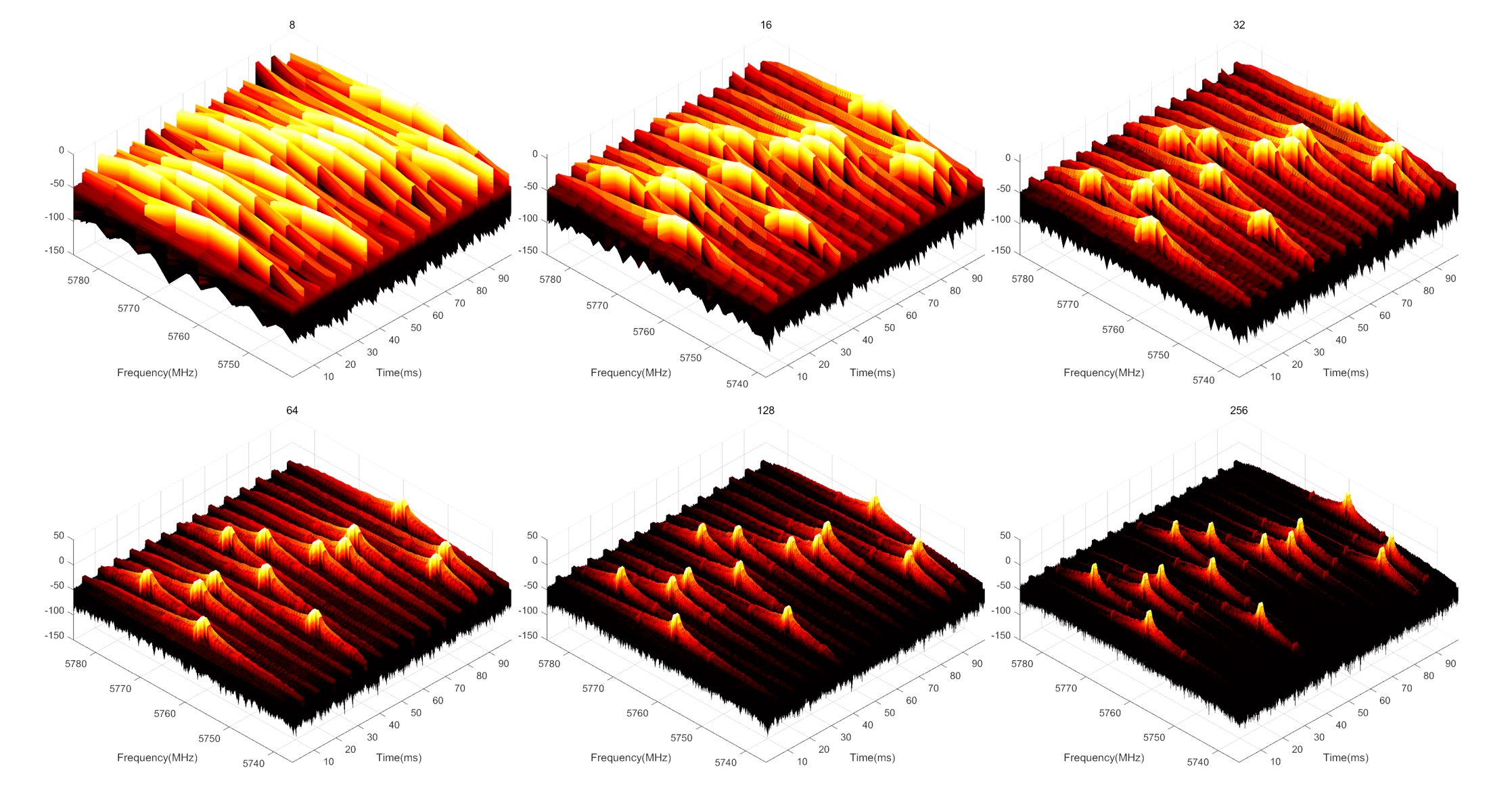}
    \caption{
    WFLY ET16S 3D spectrograms under different STFTPs %
    }
    \label{fig.app2_2}
\end{figure}

\begin{figure}[h]
    \centering\small
    \includegraphics[width=1.0\columnwidth]{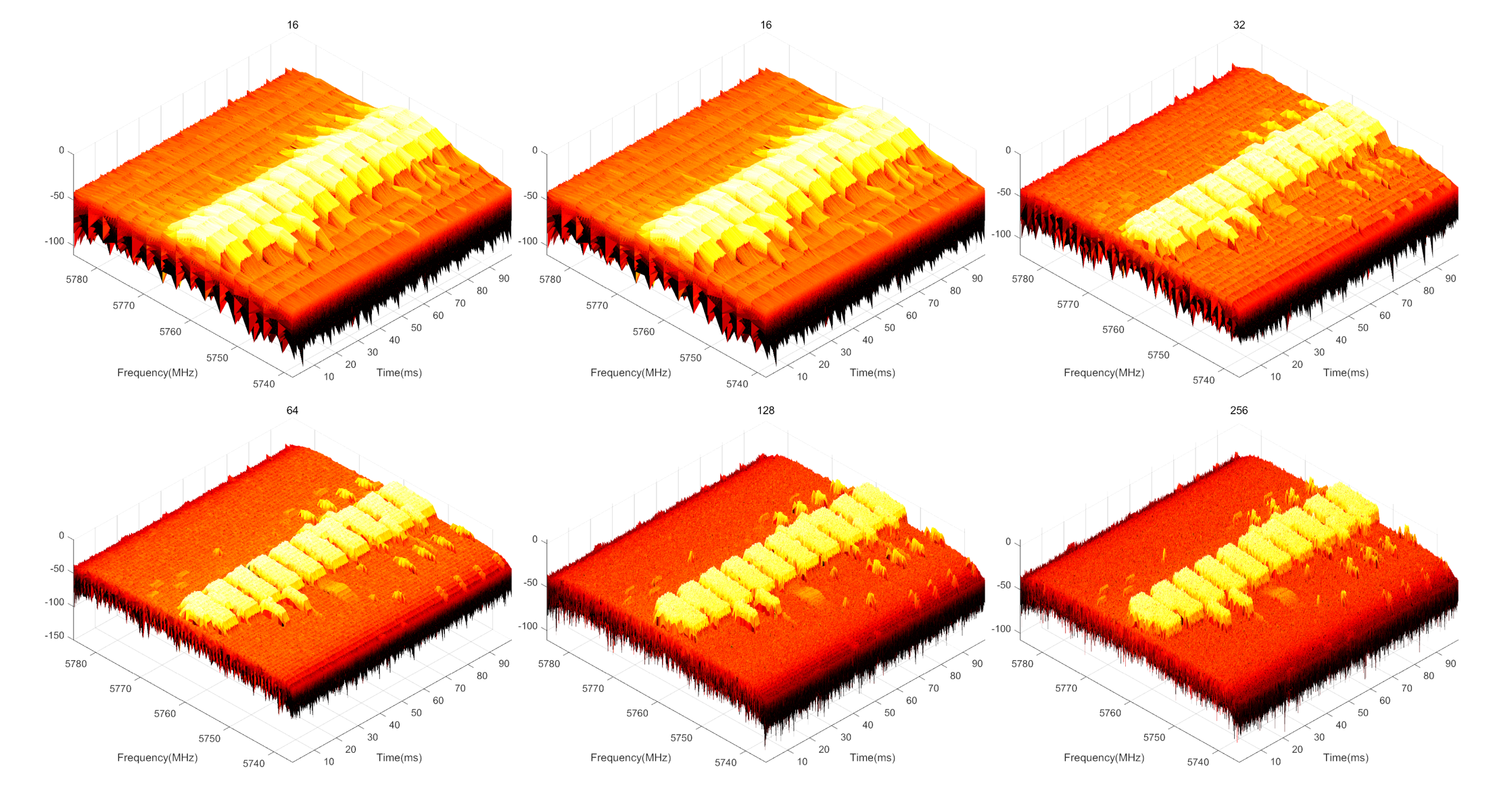}
    \caption{
    YunZhuo-h16 3D spectrograms under different STFTPs %
    }
    \label{fig.app2_3}
\end{figure}

From Eq.~\ref{eq.STFT} and Eq.~\ref{eq.hanning}, it can be observed that increasing STFTP results in a higher number of signal samples used for the STFT transformation. This leads to a richer frequency representation. To provide more detailed frequency information in the transformed signal, thereby making the frequency expression of the signal clearer and better reflecting the true frequency characteristics of the signal, we present a 3D spectrogram analysis. We select nine window lengths of \(2^n\) (where \(n = 3, 4, ..., 11\)). The results are shown in Fig.~\ref{fig.app2_1},~\ref{fig.app2_2}, and~\ref{fig.app2_3}.

As illustrated in the figure, a smaller STFTP results in a loss of frequency details in the spectrogram. When STFTP is 8, the frequency resolution of the spectrogram is limited to 12.5 MHz, meaning that frequency variations within each 12.5 MHz bandwidth cannot be captured and are represented by a single point. In this setting, only the presence of the image transmission signal at the center frequency is visible, but its boundary is unclear, and the FHSS signal cannot be distinguished. As the STFTP increases to 64, the frequency resolution improves to 1.5625 MHz, allowing the boundary of the 20 MHz bandwidth image transmission signal to be identified, although the FHSS signal remains blurred. When the STFTP reaches 512, the frequency resolution increases to 195.3125 kHz, enabling clear differentiation between the image transmission and FHSS signals. For the spectrogram of the flight control signal from the WFLY ET16S in Fig.~\ref{fig.app2_2}, when the STFTP is below 64 (i.e., the frequency resolution is below 1.5625 MHz), the broader spectrum components are indistinguishable, and the FHSS signal spectrum is barely discernible. However, once the STFTP increases to 256 and the frequency resolution reaches 390.625 kHz, the spectrum of the FHSS becomes much clearer.

Combining the experiment results presented in Sec. 5.2 with Fig. 7 (b), we can conclude that excessively high-frequency resolution is unnecessary and may excessively compromise time resolution. As shown in Fig.~\ref{fig.app2_1}, when the STFTP reaches 65536, the temporal variations in the FHSS signals begin to blend together. At an STFTP value of 524288, the image transmission signals merge entirely in the time domain, becoming indistinguishable.

This highlights the importance of selecting an appropriate STFTP synchronization to balance spectrogram quality in both time and frequency domains. If the STFTP value is either large or too small, the signal may become indistinguishable either in time or frequency.

\section{Data Collection Detail}
\label{sec:data_collection_detail}
In this subsection, we present detailed information on the data collection process to enable readers to reproduce similar procedures.

During the collection process, the transmitters of 37 different UAV types are either directly connected to the receiver of the USRPI or positioned close proximity to it, ensuring optimal signal quality of the collected data. The open-source software GNURadio is utilized to configure the acquisition parameters of the USRPI. According to previous research, the communication bandwidth used by civilian drones does not exceed 100 MHz. Therefore, a uniform sampling rate of 100 MHz is set during the data acquisition process. The center frequency of the acquisition is dynamically adjusted based on the specific signals of different drones, ensuring complete capture of the drone signals. Additionally, the gain settings are dynamically adjusted according to the SNR of the original signals to maintain a high SNR for all acquired data. Some DJI drones exhibit variability in their video transmission signal bandwidth, alternating among several predefined bandwidth options. To ensure comprehensive data coverage, multiple acquisition attempts are conducted to capture all potential scenarios, guaranteeing that each specific bandwidth configuration is represented in the dataset.

The primary challenge encountered during data collection is the uncertainty regarding the frequency bands of certain drones. To overcome this issue, we employ a spectrogram to manually search for signals across a wide frequency range. Capturing such signals incurs extremely manual costs; however, these efforts have successfully brought these raw drone signals to public attention for the first time, thereby greatly enhancing the value and relevance of the dataset.

Once the signal collection for a specific UAV is completed, we assess the usability of the acquired data using USRPO. A dedicated signal playback module, implemented using GNU Radio, enables USRPO to transmit the collected data packets. USRPO is directly connected to a spectrum analyzer, which facilitates the evaluation  of each raw data packet's bandwidth, SNR, and other pertinent characteristics. This systematic workflow for data collection ensures the quality of every signal and establishes a robust foundation for subsequent research.

\section{Dataset Details}
\label{sec:dataset_details}
In this section, we display all the drone properties in the dataset in Table.~\ref{table.dataset analysis}.

\begin{table}
\begin{center}
\caption{Summary of fundamental characteristics of drone signals in the dataset. The figure presents key statistical attributes, including file size, signal-to-noise ratio (SNR), and center frequency, frequency-hopping signal bandwidth (FHSBW, MHz), image transmission signal bandwidth (VTSBW, MHz), frequency-hopping signal duration time (FHSDT, ms), frequency-hopping duty cycle (FHSDC, ms) and frequency-hopping signal periodicity (FHSPP, ms). providing a comprehensive overview of the collected RF signal data for UAV identification and analysis."}
\label{table.dataset analysis}
\scalebox{0.9}{
\begin{tabular}{c|cccccccc}
\hline
\makecell{Type of UAV}  & \makecell{FHS BW \\ (MHz)} & \makecell{VTS BW \\ (MHz)} & \makecell{FHSDT \\ (ms)} & \makecell{FHSDC \\ (ms)} & \makecell{FHSPP \\ (ms)} & \makecell{File Size \\ (GB)} & \makecell{SNR \\ (dB)} & \makecell{MF \\ (GHz)} \\
\hline
DAUTEL EVO NANO & 4 & 20 & 4  & 29.53 & - & 29.53 & 23 & 5.77 \\
FUTABA T14SG & 32 &- & 2.3 & 30.1 & 164.1 & 53.83 & 33 & 2.44 \\

DEVENTION DEVO & 3 & - & 1 & 3 & 30 & 23.29 & 23 & 2.44 \\
Herelink HX4 & 2.96  & 19.136 & 0.52 & 5.16 & 10.09 & 19.03 & 16 & 2.42 \\
DJI FPV COMBO & 5 & 10 & 0.64 & 4 & 38.3 & 124.61 & 41 & 5.76 \\
JR PROPO XG7 & 8.402 & - & 2.17 & 12.84 & 338.67 & 16.07 & 37 & 2.44 \\
DJI AVTA2 & 7 & 10 & 0.41 & 2 & 20.6 & 99.36 & 10 & 5.77 \\
JUMPER T14 & 8.09  & - & 10.73 & 20.14 & 480 & 18.66 & 47 & 2.44 \\
DJI MAVIC3 PRO & 4.78 & 10 & 1.7 & 6 & 60 & 57.49 & 17 & 5.8 \\
JUMPER TProV2 & 4.59 & - & 6.75 & 8.99 & 35.96 & 18.53 & 32 & 2.44 \\
DJI MINI3 & 3.5 & 10 & 0.56 & 5.96 & 40.01 & 127.71 & 43 & 2.47 \\
PARROT & 18.36 & 20.01 & 0.8 & 21.5 & 21.5 & 23.68 & 19 & 5.75 \\
DJI MINI4 PRO & 6.54 & 10 & 0.404 & 2.5 & 24.048 & 47.4 & 17 & 2.45 \\
RadioMaster BOXER & 6.95 & - & 6.84 & 18.8 & 422.8 & 18.18 & 48 & 2.44 \\ 
FLYSKY EL 18 & 3 & - & 0.443 & 4.05 & 60 & 30.8 & 30 & 2.44 \\
RadioMaster TX16S & 4.59 & - & 9.3 & 19.96 & - & 15.71 & 38 & 2.44 \\
FLYSKY FS I6X & 3 & - & 1.32 & 3.82 & 61 & 28.85 & 18 & 2.44 \\
Radiolink AT9S Pro & 2.9 & - & 8.71 & 24 & - & 19.43 & 36 & 2.44 \\
FLYSKY NV 14 & 3 & - & 1.32 & 7.68 & 61.2 & 23.78 & 33 & 2.44 \\
Radiolink AT10 II & 3.125 & - & 8.81 & 24 & - & 23.58 & 33 & 2.44 \\
FRSKY X9DP2019 & 6.15 & - & 4.26 & 9 & 200 & 24.4 & 46 & 2.44 \\
SIYI FT24 & 4.59 & - & 1.61 & 12.09 & 50.05 & 25.94 & 31 & 2.44 \\
FRSKY X14 & 6.14 & - & 4.25 & 9.04 & 26.97 & 19.11 & 46 & 2.44 \\
SIYI MK15 & 3.32 & - & 1.55 & 4.985 & - & 56.72 & 41 & 2.44 \\
FRSKY X20R & 6.16 & - & 4.27 & 7.11 & 81 & 19.07 & 29 & 2.44 \\
SIYI MK32 & 7.23 & - & 0.47 & 5.01 & - & 63.98 & 41 & 2.44 \\
FUTABA T10J & 4.8 & - &1.51 & 15.07 & 466.3 & 33.7 & 37 & 2.44 \\
SKYDROID H12 & 6.06 & - & 0.25 & 2.969 & 14.381 & 27.38 & 31 & 2.47 \\
FUTABA T16IZ & 6.73 & - & 1.44 & 15.07 & 330.1  & 42.49 & 31 & 2.44 \\
SKYDROID T10 &5.273 & - & 0.27 & 2.911 & 14.722 & 42.13 & 29 & 2.47 \\
FUTABA T18SZ & 5.664 & - & 1.45 & 14.99 & 330 &  38.31 &  38 & 2.44 \\
WFLY ET10 & 4.88 & - &  0.745 & 14.3 & - & 35.32 & 31 & 2.44 \\
WFLY ET16S & 35.12 &  - & 0.752 & 3.599 & 14.5 & 24.76 & 31 & 2.44 \\
WFLY WFT09SII & 4.882 &  - & 3.13 & 1.86 & 6.81 & 29.88 & 31 & 2.41 \\
YunZhuo H12 & 4.68 & - & 2.58 & 3 & 15.3 & 23.1 & 29 & 2.46 \\
YunZhuo H16 & 10.05 & 20 & 1.05 & 5.09 & 10.19 & 45.75 & 22 & 2.46 \\
YunZhuo H30 & 4.5 & - & 1.36 & 5 & 5 & 32.14 & 22 & 2.44 \\
\hline
\end{tabular} 
}
\end{center}
\end{table}

\section{Signal Separation and Parameter Estimation Detail}\label{app3}
\begin{figure}[h]
    \centering\small
    \includegraphics[width=1.0\columnwidth]{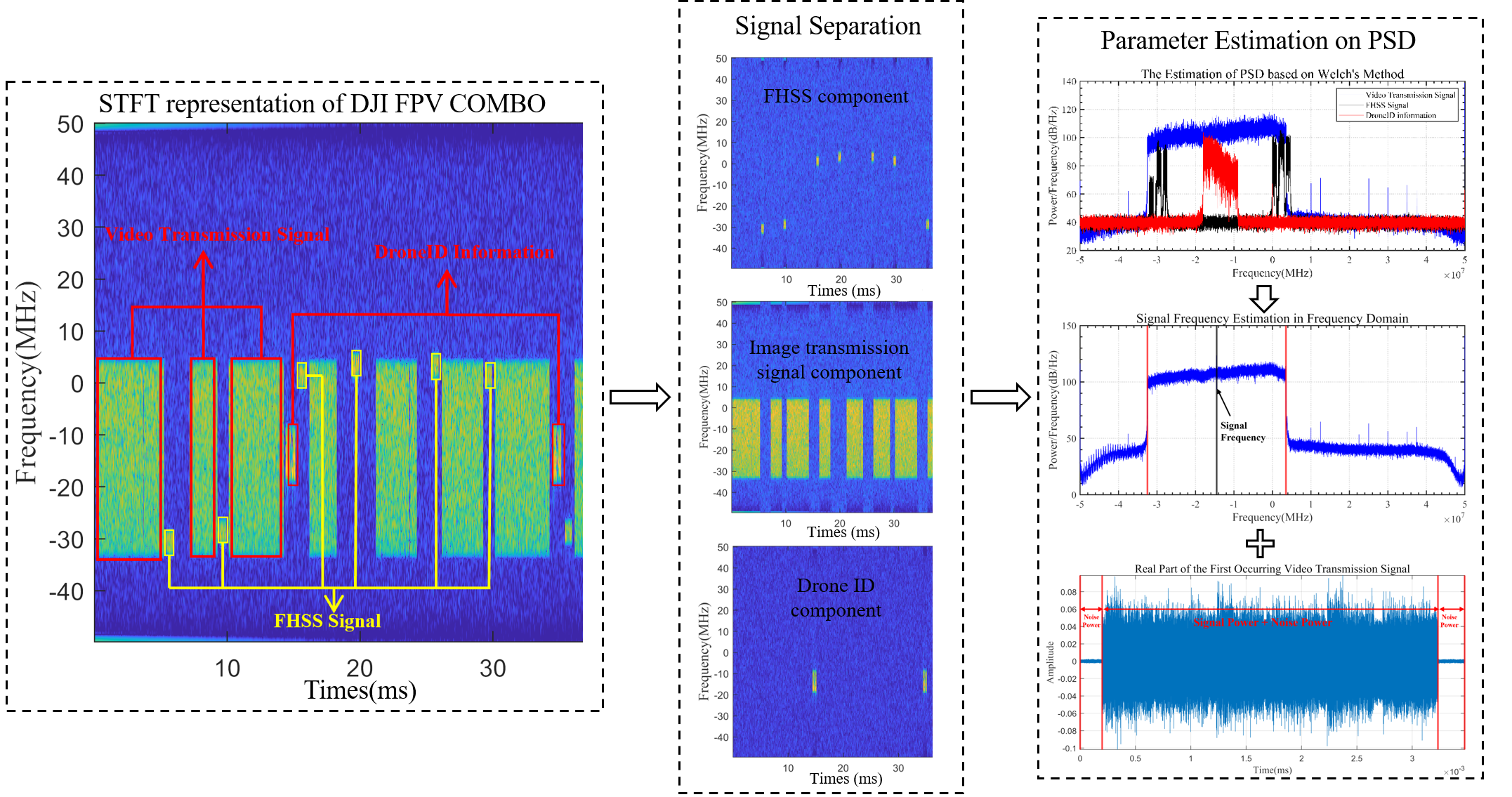}
    \caption{
    Signal analysis results %
    }
    \label{signal1}
\end{figure}

In this section, we present a signal separation and SNR estimation method using some signal processing techniques. we use an 80 ms segment of the DJI FPV COMBO radio frequency signal clip as a sample to showcase the whole process (Fig.~\ref{signal1}).

Firstly, we employ the STFT to analyze its distribution in both time and frequency domains. The window function employed in this analysis is the Hamming window, which set a length of 512. the signal sample comprises: (i) a downlink video transmission signal modulated by orthogonal frequency division multiplexing (OFDM), with a bandwidth of 40 MHz; (ii) an uplink control signal using frequency-hopping spread spectrum (FHSS) technology, characterized by a bandwidth of 3 MHz for each hop; and (iii) a data frame containing DroneID information as per the DJI OcuSync protocol, with a bandwidth of 10 MHz. Although the frequency bands employed by these three types of signals overlap, they do not coincide temporally. Consequently, we can implement a signal detection algorithm based on dual sliding windows in the time domain to identify both the starting and ending positions of the signal frames. Given that the frame durations for these three types of signals differ significantly, we can classify them based on this distinguishing characteristic. Finally, we perform STFT separately for each classification result to obtain each component.

Then, We employ Welch’s method to estimate the PSD of the signals. Welch’s method partitions the classified signal segments into multiple overlapping sections. After applying a Hamming window function to each section, the periodogram of each section is computed. Finally, we calculate the sum of all periodograms and take their average value.

Based on the separation results and prior knowledge, we take the first occurring video transmission signal (denoted as $X_{1}(n)$), and set a sliding window with a width of 40MHz on its PSD. Starting from the minimum frequency to the maximum frequency, and calculating the energy value within the window. When the energy within the window is at its maximum, we determine that the current signal matches the window function, and the frequency corresponding to the center of the window function on the frequency axis is then considered as the frequency of the signal.

Based on the starting and ending positions of the signal, we can calculate the energy of the signal and the noise in the time domain, and then estimate the SNR. Assuming the starting index of $X_{1}(n)$ is ${N_{s}}$ and the ending index is ${N_{e}}$, the length of the {$X_{1}(n)$} is $L$. Then, the sum of the signal and noise power can be expressed as:

\begin{equation}
    P_{S+N}=\frac{1}{N_e-N_s}\sum_{i=N_s}^{N_e}{\left| x\left( i \right) \right|^2}
\end{equation}
And the power of noise is:

\begin{equation}
    P_N=\frac{\sum_{i=1}^{N_s}{\left| X_1\left( i \right) \right|^2}+\sum_{i=N_e}^N{\left| X_1\left( i \right) \right|^2}}{N_s+N-N_e}
\end{equation}
Then, the estiamtion of SNR is:

\begin{equation}
    \mathrm{SNR}=10\lg \left( \frac{P_{S+N}-P_N}{P_N} \right)
\end{equation}

\section{Color Map Detail}\label{app1}
\begin{figure}[h]
    \centering\small
    \includegraphics[width=1.0\columnwidth]{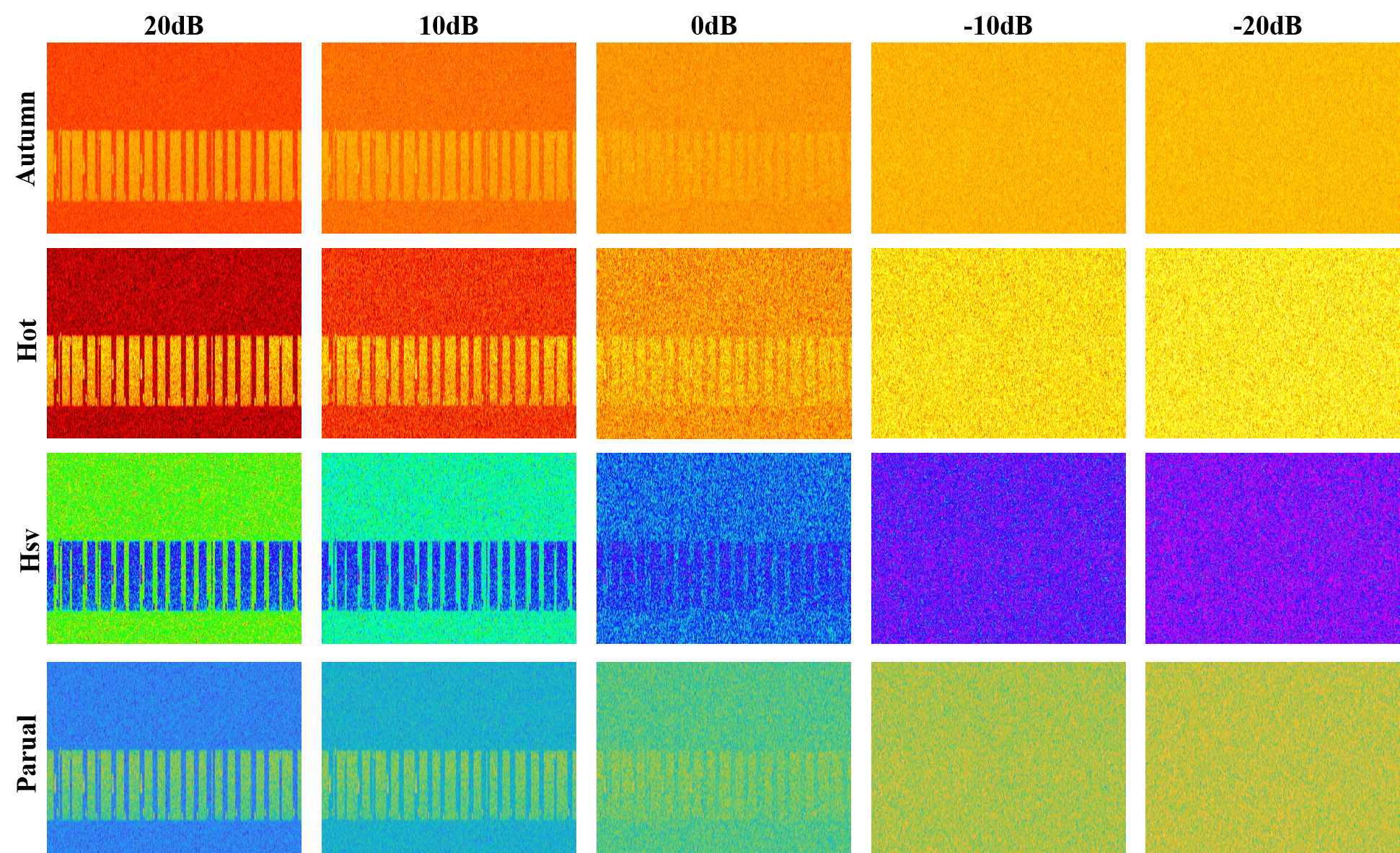}
    \caption{
    Spectrograms of DJI AVATA2 generated under different SNRs and color mapping methods %
    }
    \label{SNR}
\end{figure}

In this section, we examine the impact of different color mapping methods and SNR levels on the quality of time-frequency spectrograms. Using the dataset from the experimental subsection, we generate spectrograms for the raw signals of the DJI FPV COMBO drone at SNR levels of 20 dB, 10 dB, 0 dB, -10 dB, and -20 dB. These spectrograms are visualized using 4 types of color map methods: the "Autumn" method, which colors the STFT results from RGB (255, 0, 0) to RGB (255, 255, 0); the "Hot" method, which colors the STFT results from RGB (0, 0, 0) to RGB (255, 255, 255); the "HSV" method, which colors the STFT results from RGB (255, 0, 0) to RGB (255, 0, 255); and the "Parula" method, which colors the STFT results from RGB (62, 38, 168) to RGB (255, 255, 0). These visualizations are illustrated in Fig.~\ref{SNR}.

As shown in the figure, as the SNR decreases, drone signals are gradually overwhelmed by background noise. When the SNR drops to approximately -10 dB, drone signals become almost indiscernible in the spectrogram. Color mapping methods also significantly affect the visual identification of drone signals, as the contrast differences caused by various mapping methods can be substantial. Based on the contrast differences observed in the comparative images, the ranking is as follows: "Parula" \(<\) "Autumn" \(<\) "Hot" \(<\) "HSV".

The experimental results shown in Sec. 5.1 indicate that color mapping methods with higher contrast differences tend to rely more heavily on abundant training data and exhibit poorer generalization performance.

\begin{figure}[h]
    \centering\small
    \includegraphics[width=1.0\columnwidth]{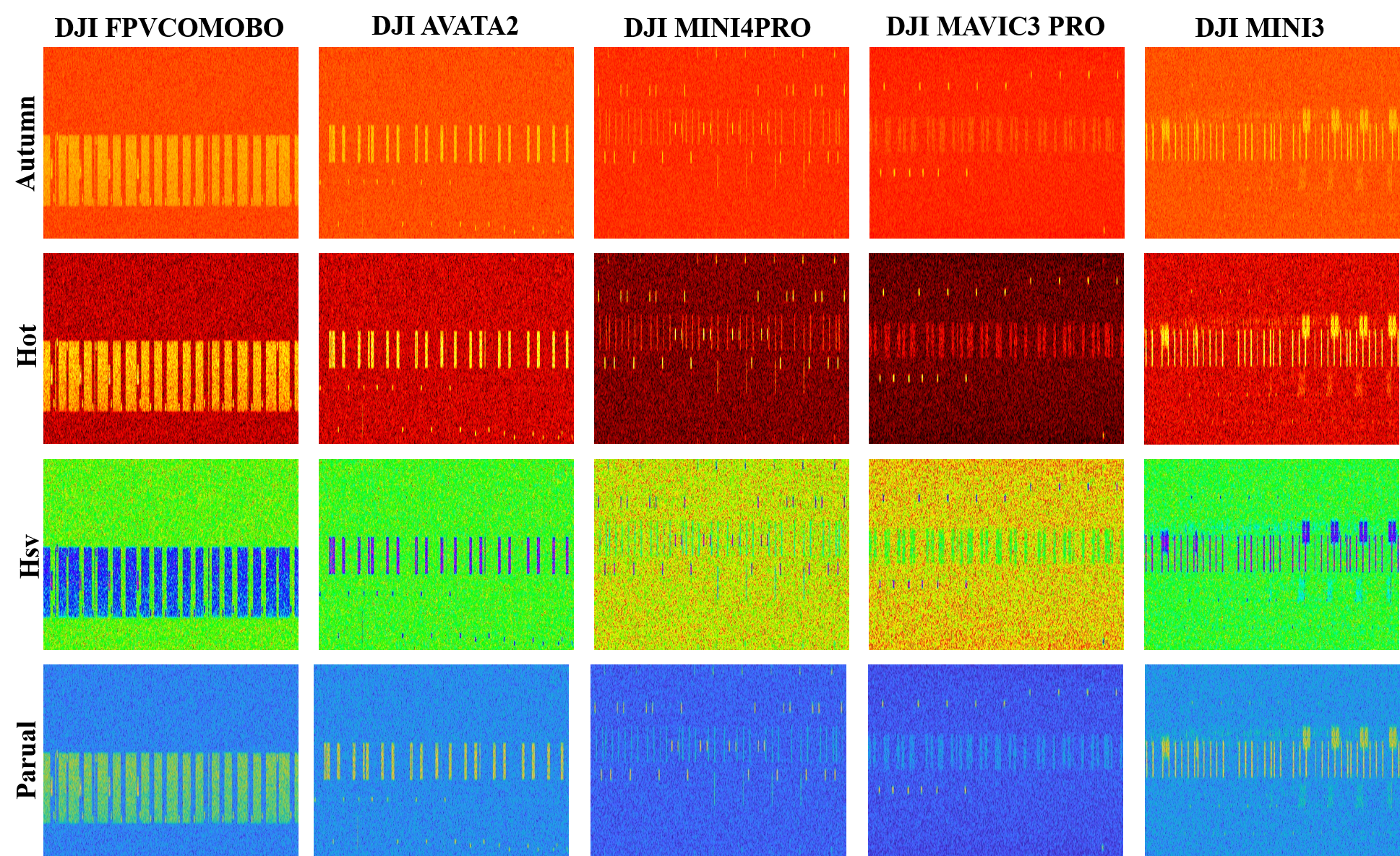 }
    \caption{
    Spectrograms of different drones generated under different SNRs and color mapping methods %
    }
    \label{DDrones}
\end{figure}

To provide readers with a more intuitive understanding of the differences between drones and their relationship with various color mapping methods, we also use the five drones mentioned in Sec. 5.1 as examples. Their spectrograms under the same color mapping methods aforementioned are illustrated in Fig.\ref{DDrones}.

As shown in the figure, it is evident that the spectrograms from different UAVs exhibit significant visual differences. Even the individual image transmission and flight control signals show notable variations.

\end{document}